\documentclass[runningheads]{llncs}

\usepackage[T1]{fontenc}
\usepackage{graphicx}
\usepackage{amsmath}
\usepackage{mathtools}

\DeclareMathOperator*{\argmin}{arg\,min}
\usepackage{amssymb}
\DeclareMathOperator*{\mean}{mean}
\usepackage{booktabs}
\usepackage{hyperref}

\newcommand{\SE}{\mathrm{SE}}
\newcommand{\E}{\mathrm{E}}
\renewcommand{\O}{\mathrm{O}}
\newcommand{\SO}{\mathrm{SO}}
\newcommand{\V}{\mathcal{V}}
\newcommand{\real}{\mathbb{R}}
\renewcommand{\P}{\mathcal{P}}
\makeatletter
\newcommand{\longdash}[1][2em]{\makebox[#1]{$\m@th\smash-\mkern-7mu\cleaders\hbox{$\mkern-2mu\smash-\mkern-2mu$}\hfill\mkern-7mu\smash-$}}
\makeatother
\newcommand{\omitskip}{\kern-\arraycolsep}
\newcommand{\llongdash}[1][2em]{\longdash[#1]\omitskip}
\newcommand{\rlongdash}[1][2em]{\omitskip\longdash[#1]}
\newcommand{\norm}[1]{\Vert #1 \Vert}
\newcommand{\T}{\mathsf{T}}
\newcommand{\plusplus}{\texttt{++} }  
\newcommand{\N}{\mathcal{N}}
\newcommand{\abs}[1]{| #1 |}
\renewcommand{\L}{\mathrm{L}}

\usepackage{xcolor}
\newcommand{\revision}[1]{\textcolor{black}{#1}}

\begin{document}

\title{$\SE(3)$ symmetry lets graph neural networks learn arterial velocity estimation from small datasets}
\titlerunning{$\SE(3)$-equivariant velocity field estimation}

\author{Julian Suk\orcidID{0000-0003-0729-047X} \and\\ Christoph Brune\orcidID{0000-0003-0145-5069} \and\\ Jelmer M. Wolterink\orcidID{0000-0001-5505-475X}}

\authorrunning{J. Suk et al.}

\institute{Department of Applied Mathematics \& Technical Medical Center,\\ University of Twente, The Netherlands\\
\email{\{j.m.suk, c.brune, j.m.wolterink\}@utwente.nl}}

\maketitle

\begin{abstract}
Hemodynamic velocity fields in coronary arteries could be the basis of valuable biomarkers for diagnosis, prognosis and treatment planning in cardiovascular disease. Velocity fields are typically obtained from patient-specific 3D artery models via computational fluid dynamics (CFD). However, CFD simulation requires meticulous setup by experts and is time-intensive, which hinders large-scale acceptance in clinical practice. To address this, we propose graph neural networks (GNN) as an efficient black-box surrogate method to estimate 3D velocity fields mapped to the vertices of tetrahedral meshes of the artery lumen. We train these GNNs on synthetic artery models and CFD-based ground truth velocity fields. Once the GNN is trained, velocity estimates in a new and unseen artery can be obtained with 36-fold speed-up compared to CFD. We demonstrate how to construct an $\SE(3)$-equivariant GNN that is independent of the spatial orientation of the input mesh and show how this reduces the necessary amount of training data compared to a baseline neural network.

\keywords{Geometric deep learning \and Graph neural networks \and Coronary arteries \and Hemodynamics.}
\end{abstract}

\begin{figure}
	\includegraphics[width=\textwidth]{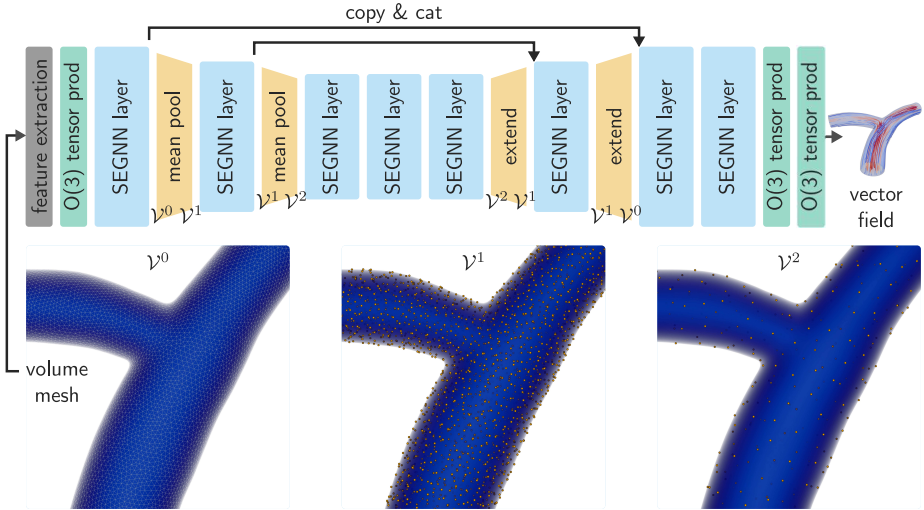}
	\caption{\textbf{Steerable $\E(3)$-equivariant graph neural network} (SEGNN). Inputs are processed by $\O(3)$ tensor product layers and SEGNN ResNet layers. We introduce mean pooling and "copy back" extension for coarsening and refinement between $\V^0$, $\V^1$ and $\V^2$. We include skip connections inside the same pooling scales. SEGNN predicts one velocity vector per mesh vertex (visualised via the resulting streamlines).}
	\label{fig:segnn}
\end{figure}

\section{Introduction}
Patient-specific, hemodynamic biomarkers have great potential in diagnosis~\cite{CandrevaNisco2022},
prognosis~\cite{BarralElSanharawi2021}
and treatment planning for patients with cardiovascular disease~\cite{ChungCebral2015}. Most studies in this field have focussed on hemodynamic biomarkers such as wall shear stress or fractional flow reserve but the post-operative change in hemodynamic velocity, a direct indicator for assessment of coronary artery bypass grafting surgery~\cite{AminWerner2018},
is also a useful biomarker. Velocity field estimates can also be used in applications like cardiovascular stent design which is an active area of medical research~\cite{BeierOrmiston2016}.
All these quantities can be accurately estimated by computational fluid dynamics (CFD), based on patient-specific 3D artery models from medical images such as MRI and CT. However, high-fidelity blood flow simulations are computationally intensive.

Recent work has demonstrated the potential for machine learning in cardiovascular biomechanics modelling~\cite{ArzaniWang2022}. In contrast to time-intensive CFD simulations, machine learning estimates can be obtained in a matter of seconds.
Estimating hemodynamic scalar and vector fields with deep neural networks, either on the artery wall by projecting it to a 1D or 2D domain and using multilayer perceptrons (MLP) and convolutional neural networks~\cite{ItuRapaka2016,SuZhang2020,GharleghiSowmya2022,FerdianDubowitz2022}, using autoencoders~\cite{LiangLiu2018,LiangMao2020} or point cloud and mesh-based methods in 3D~\cite{LiWang2021,LiSong2021,MoralesFerezMill2021,SukHaan2022} has been an ongoing area of research. In the above works, neural networks are trained offline on a dataset of results from CFD simulation which typically comprise the velocity field of the blood flow mapped to the vertices of a tetrahedral volume mesh of the artery lumen. However, only in \cite{LiangMao2020,LiSong2021,LiWang2021} a complete 3D velocity field is estimated. Other efforts have been made to learn dense velocity fields in arteries with MLPs~\cite{RaissiYazdani2020,ArzaniWang2021}, so-called physics-informed neural networks (PINN).
However, PINNs are usually limited to fitting flow patterns in a single artery and cannot generalise to other arteries, due to their use of non-local MLPs. We propose to use the generalisation capabilities of graph neural networks (GNN) to learn the relation between artery geometry and velocity field with the ability to generate accurate predictions for new and unseen arteries.

In previous work~\cite{SukHaan2022}, we have exploited gauge symmetry to estimate hemodynamic fields on the artery \textit{wall}. Here, we aim to estimate volumetric velocity fields within the artery \textit{lumen}.
We propose to make use of recent advances in $\E(3)$-equivariant message passing~\cite{BrandstetterHesselink2022}
to construct a GNN that is independent of the spatial orientation of the input geometry. Thus, our neural network is able to focus on learning the relation between geometry and hemodynamics without relying on the ambient coordinate system and to make efficient use of available training data. This is important because clinical trials in medical research typically encompass only few patients. We train and validate our GNN on a dataset of synthetic coronary bifurcations with ground truth from CFD and find that the proposed method is both more accurate and more data-efficient than a baseline PointNet\plusplus\cite{QiYi2017}.

\section{Data}
We consider the estimation of 3D velocity vectors in bifurcating coronary arteries. We synthesized 2,000 arteries after measurement statistics of left main coronary bifurcations~\cite{MedranoGraciaOrmiston2016}.
Each artery mesh consists of ca. 175,000 vertices and ca. 1,02 million tetrahedra. Steady-flow CFD was performed with SimVascular to obtain the ground truth, details of which can be found in \cite{SukHaan2022}. Each simulation took ca 15 min. The Reynolds number was low suggesting laminar flow. The dataset comprises 39 GB of simulation data and is made available upon reasonable request.

\section{Method}

\subsection{Group symmetry}
Since hemodynamics in coronary arteries are usually modelled as independent of gravity, they exhibit certain symmetries: rotation or translation of the artery in 3D does not change the flow pattern but only rotates the velocity vectors. This property is formally called equivariance under $\SE(3)$ transformation. $\SE(3)$ is the (special Euclidean) symmetry group comprised of all rotations and translations in 3D Euclidean space. In contrast, the Euclidean group $\E(3)$ consists of rotations, translations and reflections. The orthogonal group $\O(3)$ contains all distance-preserving transformations and the special orthogonal group $\SO(3)$ all rotations. A group-equivariant learning setup must be composed, from input to output, entirely of group-equivariant operators.

\subsection{Input features}
We describe the geometry of an artery by assigning a feature vector to each vertex $\V$ of the input mesh. This feature vector depends on the location of the vertex relative to the artery inlet, wall and outlets used in fluid simulation. Denote the sets of vertices comprising the artery inlet as $\P_\text{inlet}$, the artery wall as $\P_\text{wall}$ and the artery outlets as $\P_\text{outlets}$. We can construct an $\SE(3)$-equivariant descriptor of global geometry for each vertex $p^i$ as
\begin{align*}
	\kappa_\P(p) &\coloneqq \argmin_{q \in \P} \norm{p - q}_2\\
	f^i &= \left( \kappa_{\P_\text{inlet}}(p^i) - p^i,\; \kappa_{\P_\text{wall}}(p^i) - p^i,\; \kappa_{\P_\text{outlets}}(p^i) - p^i \right) \in \real^{9}\\
	i &= 1, \dots, n.
\end{align*}
We then stack these feature vectors in a vertex feature matrix
\[
X = \begin{pmatrix}
	\llongdash & f^1 & \rlongdash\\
	\llongdash & f^2 & \rlongdash\\
	& \vdots &\\
	\llongdash & f^n & \rlongdash\\
\end{pmatrix} \in \real^{n \times 9}.
\]

\begin{corollary}
	$X$ is row-wise equivariant under $\SE(3)$ transformation of $\V$.
	\label{cor:se3}
\end{corollary}
(Proof in Appendix.)

\subsection{Steerable $\E(3)$-equivariant GNN}
Steerable $\E(3)$-equivariant graph neural networks (SEGNN)~\cite{BrandstetterHesselink2022} represent signals between layers as steerable feature vectors. Steerable feature vectors consist of geometric objects, e.g. Euclidean 3-vectors, which we know how to transform with elements of the symmetry group $\E(3)$, e.g. rotations. \textbf{SEGNN layers} update the vertex features $f^i$ by message passing
\begin{align*}
m^{ij} &= \phi_m(f^i, f^j, \norm{p^j - p^i}_2^2, a^{ij})\\
f^{i} &\leftarrow \phi_f(f^i, \sum_{j \in \N(i)} m^{ij} a^i)
\end{align*}
between vertex positions $p^i$ and $p^j$ where $a^{ij}$ are edge attributes, $a^i$ are vertex attributes and $\phi_m$ and $\phi_f$ are $\O(3)$-equivariant MLPs. The MLPs are powered by the Clebsch-Gordan $\O(3)$-equivariant \textbf{tensor product}. $\N(i)$ denotes the neighbourhood of $p^i$ which we choose so that $\abs{\N(p^i)} \approx 13$. To save memory, we choose the numbers of latent channels so that SEGNN has 20,868 trainable parameters. Note that since SEGNN is $\E(3)$-equivariant, composition with our $\SE(3)$-equivariant input features creates an end-to-end $\SE(3)$-equivariant setup.

\subsubsection{Pooling}
We introduce pooling to make SEGNN efficiently pass long-range messages within large meshes. As in \cite{SukHaan2022}, we sample a hierarchy of vertex subsets $\V = \V_0 \supset \V_1 \supset \V_2$ via farthest point sampling and compute the pooling target for each $p \in \V^i$ as $\kappa_{\V^{i + 1}}(p)$.
We perform \textbf{mean pooling} among all source vertex features towards the target vertex and unpooling by "copy back" \textbf{extending} the source feature to its target vertices, unchanged. Fig.~\ref{fig:segnn} gives an overview of our neural network.

\subsection{PointNet\plusplus}
As a baseline for comparison, we implement a PointNet\plusplus\cite{QiYi2017} with five sample \& grouping layers using point convolution to pool the input down to a coarse point cloud, followed by five MLP-enhanced interpolation layers unpooling the signal up to its original size. Sampling ratios and grouping radii are chosen so that vertices have thirteen neighbours on average and point convolutions and MLPs are set up so that the overall neural network has 1,029,775 trainable parameters.

\begin{figure}[t]
	\includegraphics[width=\textwidth]{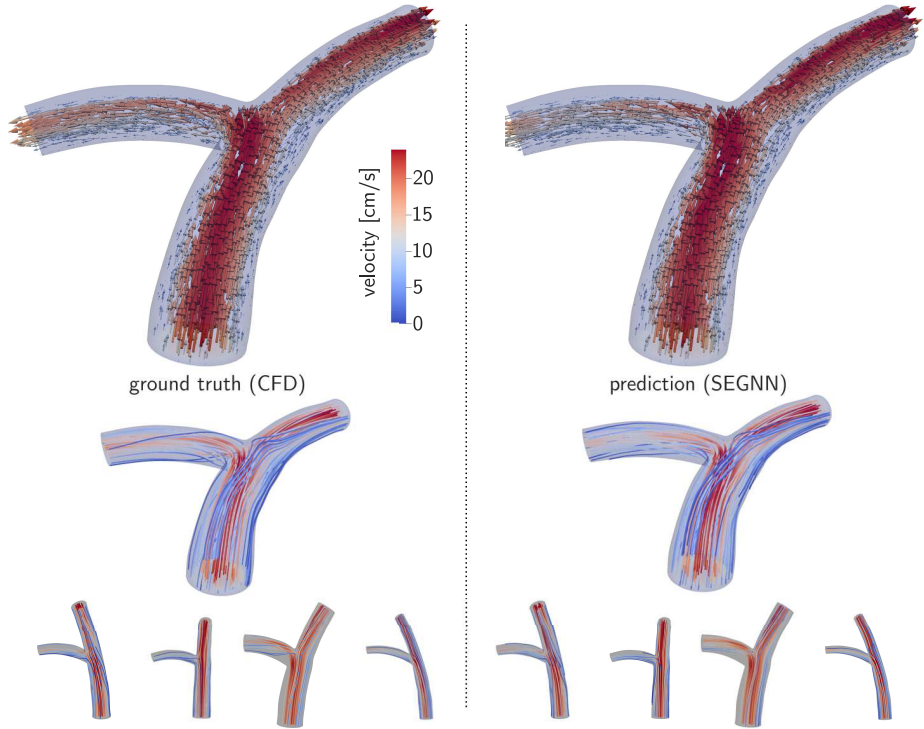}
	\caption{\textbf{Velocity field estimation.} CFD ground truth (left) and SEGNN prediction (right). On top we show a subset of the velocity vectors and on bottom we visualise the flow via a selection of streamlines evolving from the bifurcation region. We show four additional arteries of the test split.}
	\label{fig:samples}
\end{figure}

\begin{table}[t]
	\begin{center}
		\caption{\textbf{Quantitative evaluation} of velocity field estimation in the bifurcating arteries. We show mean $\pm$ standard deviation of normalised mean absolute error, approximation error $\varepsilon$ and cosine similarity $\mathfrak{cos}$ compared to ground truth CFD across the test split. Additionally, we state the number of epochs and wallclock time until convergence. PointNet\plusplus was trained and evaluated both on canonically oriented ($^\dagger$) and randomly rotated arteries ($^\ddagger$).}
		\renewcommand{\arraystretch}{1.3}
		\setlength\tabcolsep{0.2cm}
		\begin{tabular}{@{}lccccccc@{}}
			\toprule
			&& NMAE [\%] & $\varepsilon$ [\%] & $\mathfrak{cos}$ && epochs & wallclock [h:min]\\
			\midrule
			PointNet\plusplus$^\dagger$ && $1.1 \pm 0.6$ & $11.0 \pm 7.1$ & $\mathbf{0.90} \pm 0.02$ && 500 & 20:00\\
			PointNet\plusplus$^\ddagger$ && $3.4 \pm 1.5$ & $33.4 \pm 13.4$ & $0.83 \pm 0.09$ && $1000$ & 40:00\\
			SEGNN && $\mathbf{0.7} \pm 0.2$ & $\mathbf{7.4} \pm 2.2$ & $\mathbf{0.90} \pm 0.01$ && 440 & 54:54\\
			\bottomrule
			\multicolumn{8}{l}{$^\dagger$ training and test meshes are canonically oriented}\\
			\multicolumn{8}{l}{$^\ddagger$ training and test meshes are randomly rotated in 3D}
		\end{tabular}
		\label{table}
	\end{center}
\end{table}

\section{Experiments}
We divided the dataset 80:10:10 into training, validation and test split. PointNet\plusplus and SEGNN were trained with batch size two, $\L^1$ loss and Adam optimiser (learning rate $3\text{e-}4$). \revision{Systematic hyperparameter optimisation was infeasible due to long training times, so we chose hyperparameters by informed trial and error. An open-source implementation of our setup in PyG~\cite{FeyLenssen2019} and e3nn~\cite{GeigerSmidt2022}, including all hyperparameters, is available on GitHub.\footnote{\href{https://github.com/sukjulian/segnn-hemodynamics}{github.com/sukjulian/segnn-hemodynamics}}} PointNet\plusplus was trained on a single Nvidia GeForce RTX 3080 (144.10 s per epoch) while SEGNN training was accelerated with four Nvidia A40 GPUs (449.27 s per epoch). We trained each neural network until convergence, indicated by plateauing validation loss. For a new and unseen volumetric artery mesh with ca. 175,000 vertices, velocity field estimation took ca. 24.5 s of which 24 s are due to pre-processing of the input mesh and 0.5 s is due to the forward pass.

\subsection{Quantitative evaluation metrics}
We evaluate accuracy by normalised mean absolute error (NMAE), approximation error $\varepsilon$ and mean cosine similarity $\mathfrak{cos}$. Let $Y^i \in \real^{n \times 3}$ be the GNN's stacked output matrix and $V^i \in \real^{n \times 3}$ the corresponding velocity ground truth matrix for the $i$-th input mesh of the test split. We define the metrics element-wise as
\begin{align*}
\text{NMAE}_i &= \frac{\mean\limits_j \norm{(V^i - Y^i)_j}_2}{\max\limits_k\max\limits_j \norm{(V^k)_j}_2}\\
\varepsilon_i &= \frac{\sum\limits_j\norm{(V^i - Y^i)_j}_2^2}{\sum\limits_j \norm{(V^i)_j}_2^2}\\
\mathfrak{cos}_i &= \mean_j \cos(\sphericalangle (V^i)_j, (Y^i)_j),
\end{align*}
where $(\cdot)_j$ means taking the $j$-th row vector. Note that $\cos(\sphericalangle (V^i)_j, (Y^i)_j)$ is 1 if $(V^i)_j$ and $(Y^i)_j$ are proportional, 0 if they are orthogonal and -1 if they are opposing.

\subsection{Velocity field estimation}
Fig.~\ref{fig:samples} shows example results of velocity field estimation in unseen arteries. Qualitative comparison of vector field and streamlines suggest good agreement between CFD and SEGNN prediction. Even though streamlines are sensitive to small perturbations, they largely coincide. Table~\ref{table} contains quantitative evaluation showing that SEGNN strictly outperforms PointNet\texttt{++}. Both neural networks' estimation has good directional agreement to CFD, indicated by high mean cosine similarity $\mathfrak{cos}$. We train PointNet\plusplus on canonically oriented samples and evaluate its accuracy for this case, then later randomly rotate the arteries in 3D during further training and testing. In the rotated case, PointNet\plusplus is not able to recover its accuracy of the oriented case with the given input features. Note that since SEGNN is fully equivariant to rotations, its accuracy does not change for rotated input meshes.

\begin{figure}[t]
	\includegraphics[width=\textwidth]{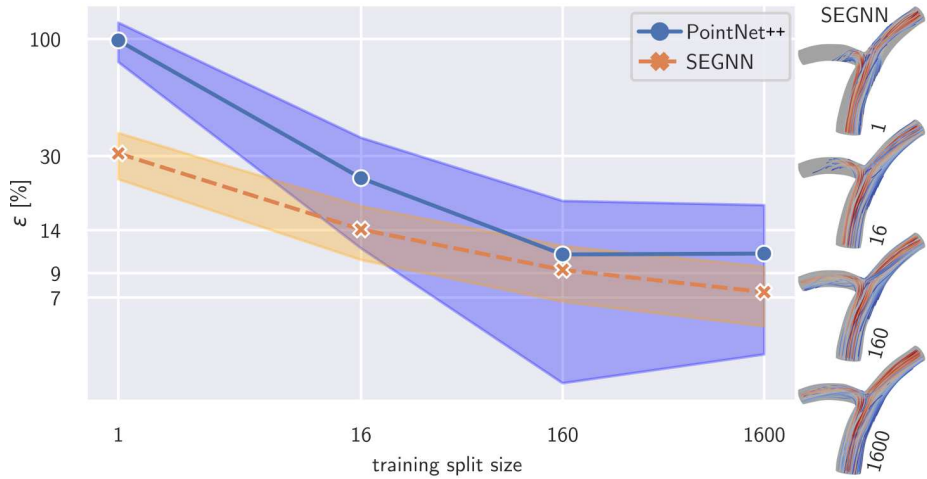}
	\caption{\textbf{$\varepsilon$ over training split size} in $\log$-$\log$ scale. Data points are the mean approximation errors and the shaded region corresponds to $\pm$ one standard deviation. The same test split (200 arteries) is used to evaluate $\varepsilon$ for each training split. Streamlines show SEGNN outputs in a test split artery after training on different split sizes.}
	\label{plot}
\end{figure}

\subsection{Learning from small datasets}
Fig.~\ref{plot} shows approximation error $\varepsilon$ resulting from PointNet\plusplus and SEGNN trained with different amounts of data \revision{in $\log$-$\log$ scale}. We find that both neural networks can obtain good accuracy when trained on 160 instead of 1,600 meshes. The performance of PointNet\plusplus decreases more rapidly than SEGNN with decreasing training split size. \revision{Training PointNet\plusplus on 160 versus 16 meshes doubles its approximation error $\varepsilon$ from $10.9 \%$ to $23.9 \%$ while SEGNN deteriorates $1.5$-fold from $9.3 \%$ to $14.1 \%$. We conclude that SEGNN is favourable for clinical trials in medical research that encompass less than 160 patients.} Intriguingly, SEGNN seems to be able to mildly learn and generalise based on training on a single artery, in contrast to PointNet\texttt{++}.

\section{Discussion and conclusion}
We demonstrate how to leverage the generalisation power of GNNs to learn to estimate 3D velocity fields in the artery lumen of the left main coronary bifurcation. Compared to CFD, this leads to a speed-up from 15 min to 24.5 s per artery. We show that our $\SE(3)$-equivariant SEGNN outperforms PointNet\plusplus in accuracy and data-efficiency. In our experiments, PointNet\plusplus is not able to accurately estimate velocity fields in rotated arteries, even when trained on rotated input meshes. We introduce pooling and parallelise over multiple GPUs to make SEGNN efficiently pass long-range messages while scaling to large input meshes.

Other works have looked into the prediction of velocity fields in arteries. Liang et al.~\cite{LiangMao2020} proposed a neural network that predicts velocity fields using autoencoder-based shape encoding of the input geometry. Their method requires vertex correspondence between input meshes, achieved trough reparametrisation, to ensure an equal number of mesh vertices with the same nodal connectivity. This can be overly restricting if input geometries differ significantly, e.g. branching artery trees. Li et al.~\cite{LiWang2021,LiSong2021} proposed an adapted PointNet~\cite{QiSu2017} that combines global and local information to predict velocity fields on arbitrary point clouds. While PointNet is robust to ordering of the input vertices due to the locality of its layers (permutation equivariance), its output implicitly depends on the ambient coordinate system in which its input points are expressed. This may not be a problem if the input geometries are all canonically oriented by construction. However, this is not common in practice and thus PointNet requires canonical alignment ("registration") of the input geometries. Furthermore, the implicit conditioning on the ambient coordinate system misguides the neural network's prediction. Addressing this by data augmentation might waste expressive capacity or may not be possible at all, as our results indicate.
In contrast, SEGNN is by design fully equivariant to roto-translations and thus robust to misaligned data. In our experiments, we find that SEGNN is able to estimate velocity fields in new and unseen arteries based on training even with small datasets. We conjecture that due to its $\SE(3)$ equivariance, SEGNN avoids conditioning on misleading information like the alignment of the input mesh, which is a key driver of performance.

Clinical datasets with patient-specific artery models are difficult to obtain and typically small. Thus, it is imperative to further increase the data efficiency of our method. To do so, we plan to incorporate knowledge about the flow physics, leading to so-called physics-informed graph networks~\cite{ShuklaXu2022}. A limitation of our study is the reliance on a dataset of synthetic coronary arteries as well as the steady-flow assumption. To address this, we aim to extend our work to real-life patient data and pulsatile flow in the future. Furthermore, we will condition our GNNs on flow boundary conditions like inflow velocity and outlet pressure.

In conclusion, SEGNN is able to learn the relation between artery geometry and hemodynamic velocity in coronary arteries based on training with small datasets while being robust to misalignment of the input meshes through $\SE(3)$ equivariance.

\subsubsection{Acknowledgements}
This work is funded in part by the 4TU Precision Medicine programme supported by High Tech for a Sustainable Future, a framework commissioned by the four Universities of Technology of the Netherlands. Jelmer M. Wolterink was supported by the NWO domain Applied and Engineering Sciences VENI grant (18192). This work made use of the Dutch national e-infrastructure with the support of the SURF Cooperative using grant no. EINF-2675.

We would like to thank Pim de Haan and Johannes Brandstetter for the fruitful discussions about steerable equivariant graph neural networks.

\bibliographystyle{splncs04}
\bibliography{references}

\section*{Appendix}

\subsection*{Proof of Corollary~\ref{cor:se3}}
\begin{proof}
	Vertex positions $p$ translate as $p + t$ where $t \in \real^3$. Since $\kappa_\P$ and $x^i$ are defined in terms of differences between vertex position, they are invariant under uniform translation of $\V$. Consider the rotation $r \in \SO(3)$ and its representation as orthogonal matrix $R \in \real^{3 \times 3}$. Then
	\[
	\kappa_{r \P}(r p) = \argmin_{q' \in r \P} \norm{R p - q'}_2 = R \kappa_\P(p)
	\]
	where $q' = R q$ and since
	\[
	\norm{R (p - q)}_2 = \sqrt{(p - q)^\T R^\T R (p - q)} = \norm{p - q}_2
	\]
	Consequently the rows of $X$ transform as
	\[
	r f^i = \left( R (\kappa_{\P_\text{inlet}}(p^i) - p^i),\; R (\kappa_{\P_\text{wall}}(p^i) - p^i),\; R (\kappa_{\P_\text{outlets}}(p^i) - p^i) \right)
	\]
	under uniform rotation of $\V$.
\end{proof}

\subsection*{Normalisation}
BatchNorm typically computes normalisation statistics channel-wise over the entire training batch (consisting of multiple graphs). Since the statistics may differ between training and testing batches, running estimates stored during training are used during testing. We propose to instead compute statistics over each disjoint graph during both training and testing. However, since this incurs additional computational cost especially over large graphs, we can use regular BatchNorm and disable "evaluation mode" as an approximation when working with small batch sizes.

\end{document}